\documentclass[letterpaper]{article} 
\usepackage{aaai25}  
\usepackage{times}  
\usepackage{helvet}  
\usepackage{courier}  
\usepackage[hyphens]{url}  
\usepackage{graphicx} 
\usepackage{natbib}  
\usepackage{caption} 

\usepackage{algorithm}
\usepackage{algorithmic}
\usepackage{times}
\usepackage{soul}
\usepackage{url}
\usepackage[utf8]{inputenc}
\usepackage{caption}
\usepackage{subcaption}
\usepackage{graphicx}
\usepackage{amsmath}
\usepackage{amsthm}
\usepackage{booktabs}
\usepackage{algorithm}
\usepackage{algorithmic}
\usepackage[switch]{lineno}
\usepackage{amssymb}
\usepackage{multirow}
\usepackage{subcaption}
\usepackage{adjustbox}

\usepackage{algorithm}
\usepackage{algorithmic}

\usepackage{newfloat}
\usepackage{listings}
\DeclareCaptionStyle{ruled}{labelfont=normalfont,labelsep=colon,strut=off} 
\floatstyle{ruled}
\newfloat{listing}{tb}{lst}{}
\floatname{listing}{Listing}
\title{Edge Contrastive Learning: An Augmentation-Free Graph Contrastive Learning Model}

\author {
    Yujun Li\textsuperscript{\rm 1},
    Hongyuan Zhang\textsuperscript{\rm 2,3},
    Yuan Yuan\textsuperscript{\rm 1}\thanks{Corresponding author.}
}
\affiliations {
    \textsuperscript{\rm 1}School of Artifcial Intelligence, Optics and Electronics (iOPEN), \\
Northwestern Polytechnical University, Xi'an 710072, P.R. China\\
    \textsuperscript{\rm 2}Institute of Artificial Intelligence (TeleAI), China Telecom, P. R. China\\
    \textsuperscript{\rm 3}The University of Hong Kong\\
    yujunli361@gmail.com, hyzhang98@gmail.com, y.yuan1.ieee@gmail.com
}

\usepackage{bibentry}

\begin{document}

\maketitle
\begin{abstract}
Graph contrastive learning (GCL) aims to learn representations from unlabeled graph data in a self-supervised manner and has developed rapidly in recent years. However, edge-level contrasts are not well explored by most existing GCL methods. Most studies in GCL only regard edges as auxiliary information while updating node features. One of the primary obstacles of edge-based GCL is the heavy computation burden. To tackle this issue, we propose a model that can efficiently learn edge features for GCL, namely {\bf{A}}ugmentation-{\bf{F}}ree {\bf{E}}dge {\bf{C}}ontrastive {\bf{L}}earning (AFECL) to achieve edge-edge contrast. AFECL depends on no augmentation consisting of two parts. Firstly, we design a novel edge feature generation method, where edge features are computed by embedding concatenation of their connected nodes. Secondly, an edge contrastive learning scheme is developed, where edges connecting the same nodes are defined as positive pairs, and other edges are defined as negative pairs. Experimental results show that compared with recent state-of-the-art GCL methods or even some supervised GNNs, AFECL achieves SOTA performance on link prediction and semi-supervised node classification of extremely scarce labels. The source code is available at https://github.com/YujunLi361/AFECL.
\end{abstract}

%

\section{Introduction}

The recent success of graph neural networks has driven research in knowledge graphs \cite{zhu2022neural}, recommendation \cite{cao2022cross,li2023making}, 
e-commerce \cite{zhang2022efraudcom}, biological systems \cite{rao2022communicative}, etc. 
A common approach to training graph neural networks is to use the supervised mode. For supervised learning, a sufficient amount of input data and label pairs are given otherwise this leads to overfitting \cite{feng2020graph}. However, the supervised training process requires a large amount of labeled data and is expensive in real applications, since labels are expensive, or even long-tailed data \cite{yang2023imbalanced}.

Unsupervised and self-supervised learning (SSL) trains neural network models on unlabeled data, reducing dependence on labeled data 
\cite{cai2021task,zhang2023non,li2023bio,zhang2024decouple}. 
Among SSL methods, contrastive learning (CL) has been shown to achieve comparable performance levels to its supervised counterparts across a spectrum of tasks, including Computer Vision (CV) \cite{chen2020simple} and Natural Language Processing (NLP) \cite{gao2021simcse}. With the development of CL, CL was introduced into the graph domain, combining GNN and CL to learn embedding, also dubbed graph contrastive learning (GCL)      \cite{hassani2020contrastive,zhu2021graph,xia2022simgrace}.

Recent studies on graph contrastive learning focus on a similar contrastive learning paradigm.
Specifically, graph contrastive learning aims to learn one or more encoders such that similar instances in the graph agree with each other, and dissimilar instances disagree with each other. Most existing GCL methods can be divided into two aspects: graph view generation by applying different transformations and learning towards minimizing contrastive loss based on mutual information (MI) estimators. 
For graph view generation, there are various graph transformation (augmentation) methods, such as node attribute masking \cite{you2020graph}), edge perturbation \cite{qiu2020gcc,you2020graph}, graph diffusion \cite{hassani2020contrastive}, $\pi$-noise augmentation \cite{li2022positive,zhang2024data}, etc. For contrastive objectives, there are contrastive losses widely used in CL, such as Donsker-Varadhan estimator \cite{donsker1983asymptotic,belghazi2018mutual}, noise contrastive estimation 
(InfoNCE) \cite{gutmann2010noise,oord2018representation}, normalized temperature-scaled cross-entropy (NT-Xent) \cite{sohn2016improved}, to estimate and maximize the mutual information in CL computationally. However, there are some drawbacks of this paradigm.


On the one hand, MAGCL \cite{gong2023ma} argues augmentations that retain sufficiently complete task-relevant information cannot generate diverse enough views. Existing handcrafted graph augmented strategies may destroy the graph structure resulting in failing to maintain the task-related information intact. At the same time, using two encoders with the same neural architecture and tied parameters for the augmented graphs may hurt the diversity of the augmented views.
Specifically, removing important edges can severely damage graph topology that is highly relevant to downstream tasks, resulting in poor graph embedding quality \cite{zhu2021graph}. Generating augmentations in a heuristic \cite{zhu2021contrastive} or adversarial \cite{you2021graph,feng2022adversarial} manner, using two view encoders with the same architecture, may also lead to graph embedding quality low.

On the other hand, most existing GCL methods directly fetch the contrastive goal originally proposed in CL \cite{qiu2020gcc,zhu2020deep,wan2021contrastive,you2021graph,zhu2021contrastive}, while paying no attention to the topological information and the information of edges in the graph data. They regard different augmentations of the same sample (node or graph) as positives and different augmentations of other samples as negatives and pull the positives close and the negatives far apart. However, they are usually based on the homophily assumption, which means that connected nodes should be more similar. In addition,  for graphs, changes in edges can directly affect topological information.
Exploiting the contrastive objective ignores topological information and contradicts the homophily assumption.

To address the above problems, we propose a new paradigm that applies edge contrastive learning without data augmentation. Unlike the existing 
augmentation-free method \cite{xiao2024simple} in node level, we are the first to propose and verify the validity of edge-level contrast called {\bf{A}}ugmentation-{\bf{F}}ree {\bf{E}}dge {\bf{C}}ontrastive {\bf{L}}earning (AFECL). Specifically, our model adopts a simple graph neural network (GNN) as the encoder to generate node embeddings. Then, instead of only treating edges as auxiliary information to implement parameter updates, we directly compute the edge embeddings through node embeddings.  For the contrastive objectives, AFECL fully considers network topological information and regards edges connecting the same node as positive pairs and edges that do not connect nodes as negative pairs. Overall, our main contributions are listed as follows:

\begin{itemize}
\item To the best of our knowledge, for the first time we study the edge-level pairs for contrast. We introduce a novel edge representation learning method generating edge embeddings through node embeddings. Moreover, we design a new edge-level contrastive loss where edges connecting the same node as positive pairs and edges that do not connect nodes as negative pairs.
\item We develop a novel and effective paradigm for graph contrastive learning which provides a high degree of flexibility, high efficiency, and ease
of use. Specifically, AFECL does not need additional handcrafted graph augmentations and  can train on
graphs at any scale by generating a small number of edge features.
\item Experimental results show that, compared with recent state-of-the-art GCL methods or even some supervised GNNs, our method achieves SOTA performance on link prediction and semi-supervised node classification of extremely scarce labels.
\end{itemize}

\section{Related Work}

In this section, we provide a brief review of existing GCL methods. Then, the proposed method and its related work are summarized.

\subsection{Data Augmentations for Contrastive Learning}

The general paradigm in CL requires data augmentation of the input data first.
In computer vision (CV), there are many high-quality data augmentation schemes, such as rotations, blurring, resizing, cropping, flipping, etc \cite{chen2020simple}.
In natural language processing (NLP), due to its discrete nature, the data augmentation scheme is not as straightforward as in CV, such as word
deletion, reordering, substitution, etc \cite{gao2021simcse}. In the graph domain, there are many data augmentation methods, but there is no universal graph data augmentation scheme. For example, DGI \cite{veličković2018deep} achieves data augmentation by disrupting the nodes of the original graph.
GraphCL \cite{you2020graph} proposes to achieve data augmentation through attribute masking, edge perturbation, node dropout and subgraph extraction.
GRACE \cite{zhu2020deep}, GCA \cite{zhu2021graph} and CCA-SSG \cite{zhang2021canonical} generate views by randomly masking node attributes and randomly removing edges. MoCL \cite{sun2021mocl}demonstrates that infusing domain knowledge can serve as prior knowledge, which helps to find appropriate augmentations. MVGRL \cite{hassani2020contrastive} augments the input graph via graph diffusion and ARIEL \cite{feng2022adversarial} augments the input graph via adversarial graph perturbation. SPAN \cite{lin2023spectral} generates augmentations by maximizing the spectral change. However, there are limitations in that appropriate augmentation methods must be selected for different graph datasets to the above handcraft augmentations, resulting in poor generalizability \cite{you2020graph}. To avoid manually tuning dataset-specific graph augmentations, LOCAL-GCL \cite{zhang2022localized} 
and GraphACL \cite{xiao2024simple} define first-order neighbors of nodes as positive samples without augmentation. 

The proposed model regards the original graph as the view, which not only avoids improper modification of the graph dataset topology but also has wide generalizability.
\begin{table}[]

\centering
    \begin{adjustbox}{width=0.47\textwidth} 
\begin{tabular}{cccc}
\toprule
Method         & Data Augmentation                        & Views      & Contrastive Objective \\ \hline
DGI            & No Data Augmentation                     & 1          & Node-graph            \\ \hline
GMI            & No Data Augmentation                     & 1          & Node-node             \\ \hline
MVGRL          & Graph Diffusion                          & 2          & Node-graph            \\ \hline
GCC            & Extract Subgraph                         & 2          & Graph-graph           \\ \hline
GCA            & Attribute Perturbation+Edge Perturbation & 2          & Node-node             \\ \hline
ARIEL          & Adversarial Graph Perturbation           & 2          & Node-node             \\ \hline
SPGCL          & No Data Augmentation & 1          & Node-node             \\ \hline
GraphACL          & No Data Augmentation                      & 2          & Node-node             \\ \hline
PiGCL          & Attribute Perturbation+Edge Perturbation                      & 2          & Node-node             \\ \hline
\textbf{AFECL} & \textbf{No Data Augmentation}            & \textbf{1} & \textbf{Edge-edge}    \\ \bottomrule
\end{tabular}}
\end{adjustbox}
\caption{Comparison of related works.}
\label{comparsion}
\end{table}
\subsection{Graph Contrastive Learning Models}

\begin{figure*}[htbp]
\centering
\includegraphics[scale=0.32]{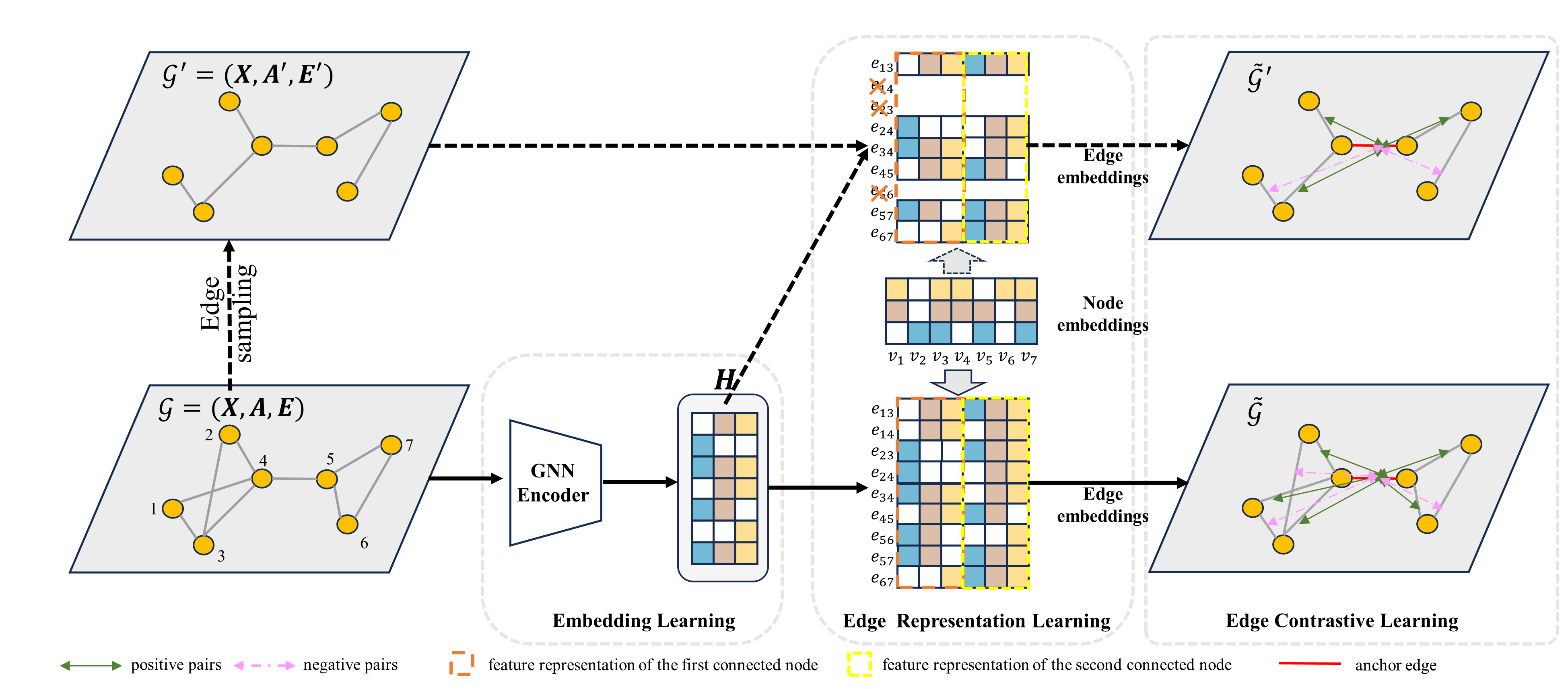}
\caption{The architecture of the proposed AFECL framework. Firstly, the original graph is regarded as a view and fed into the encoder to learn the representation of the nodes. Then, the learned node representations are used to concatenate and generate edge representations. Note that in large graphs, edges need to be sampled first to generate representations behind the sampling. In the end, the model applies edge contrastive loss to maximize the mutual information between positive pairs of representations and minimize the mutual information between negative pairs of representations. The red edges represent both anchors and positives.}
\label{figure1}
\end{figure*}
GCL often explores node-node, node-graph and graph-graph relationships as contrastive objectives \cite{xie2022self}. For node-graph contrast, most GCL methods employ graph-level encoders and node-level encoders to obtain the corresponding embeddings. For example, DGI \cite{veličković2018deep} contrasts node embeddings of the original graph and node embeddings of the corrupted graph with graph embeddings. MVGRL \cite{hassani2020contrastive} contrasts the embedding of a view through a node-level encoder with the embedding of another view through a graph-level encoder. For graph-graph contrast, most GCL methods employ two graph-level encoders to obtain graph embeddings and contrast graph-level between positives and negatives. For example, GCC \cite{qiu2020gcc} contrasts different subgraph embeddings obtained by different graph-level encoders. For node-node contrast, most GCL methods employ node-level encoders to obtain node embeddings and contrast node-level between positives and negatives. For example, GRACE \cite{zhu2020deep} and GCA \cite{zhu2021graph} contrast the embedding of each node, where the same node in different augmented views is regarded as positive pairs and other nodes are viewed as negative pairs. However, these methods do not consider the topological information of the graph. 
NCLA \cite{shen2023neighbor} proposes Neighbor Contrastive Loss, which treats nodes with different augmentations and their neighbor nodes as positive pairs, and other nodes as negative pairs. In addition, existing GCL methods ignore edge information and only use edges as auxiliary information to update node representations.

In our model, we propose a new edge contrastive loss for edge-edge GCL. Unlike existing contrastive objectives, our method is an edge-edge contrast, where edge representations are generated from node representations.
Moreover, our method exploits network topology information to define positive and negative pairs, where positives are the edges connecting the same nodes and negatives are the others.

{\textit{Comparisons with related graph CL methods.}}  In summary, we briefly compare the proposed AFECL with other proposed state-of-the-art graph contrastive learning methods, as shown in Table \ref{comparsion}. It can be seen that the proposed AFECL method requires only one view and does not require the use of additional node corruption like DGI \cite{veličković2018deep}. Besides, these data augmentation methods have been proven to be effective, but they suffer from problems such as complexity and possible inappropriate modification of the original graph.
AFECL does not require data augmentation and only uses the original graph faded into the encoder. Most importantly, AFECL is the only model that proposes to define edge-edge as the contrastive objective.

\section{Methodology}

In this section, we present AFECL in detail. Then, we elaborate on how to generate edge features in Section. 3.2, how to design the edge contrastive loss and define positives and negatives between edges in Section. 3.3. We propose a novel edge contrastive architecture for unsupervised graph representation learning, as shown in Figure \ref{figure1}.

\subsection{Preliminaries}

Let $\mathcal{G} = (\mathcal{V}, \mathcal{E})$ denote a graph, where
$\mathcal{V}= \{v_1, v_2, \cdots, v_N\}$ and $ \mathcal{E} \subseteq \mathcal{V} \times \mathcal{V}$ represent the node set
and edge set respectively. $\boldsymbol{X} \in $
$\mathbb{R}^{N \times F}$ and $\boldsymbol{A} \in \{0, 1\}^{N \times N}$ denote the node feature matrix and the symmetric adjacency matrix, 
where $\boldsymbol{x}_i \in \mathbb{R}^{F}$ is the feature vector of $v_i$ and $A_{ij}=1$ iff $(v_i, v_j) \in \mathcal{E}$, otherwise $A_{ij}=0$.
$\mathcal{N}_i$ represents the first-order neighbors of node $i$ in the graph.
$\boldsymbol{E} \in \mathbb{R}^{M \times D}$ denotes the edge feature matrix, where $\boldsymbol{e}_{ij} \in \mathbb{R}^{D}$ is the feature vector
of edge$(v_i,v_j)$ and $e_k$ represents the $k$-th edge. Given $\boldsymbol{X}$ and $\boldsymbol{A}$ as the input, the proposed model employs the GNN encoder $f(\boldsymbol{X}, \boldsymbol{A})$
to learn the representations of nodes $\boldsymbol{H} = f(\boldsymbol{X}, \boldsymbol{A}) \in \mathbb{R}^{N \times F^{\prime}}$, $F^{\prime} \ll F$.
Then, the proposed method exploits the embeddings of nodes to generate the representations of edges. These representations are learned by optimizing the
the edge contrastive loss, and can be used in downstream tasks without access to the labels. 

\subsection{Edge Representation Learning}

In our proposed model, we employ multi-head GAT as the encoder for all benchmark datasets. Each head is viewed as a single-layer
feedforward neural network. In the $k$-th head attention, the edge coefficient between neighbor node $i$ and $j$ can be expressed as

\begin{equation}
\resizebox{0.9\linewidth}{!}
{$\begin{gathered}
\displaystyle
    \alpha_{ij}^{(k)} = \frac{\exp({{\rm LeakyReLU}( \boldsymbol{a}^{(k)}[\boldsymbol{W}^{(k)} \boldsymbol{x}_i \| \boldsymbol{W}^{(k)} \boldsymbol{x}_j])})}
				{\sum_{v_p \in \mathcal{N}_i \cup \{v_i\}} \exp({{\rm LeakyReLU}( \boldsymbol{a}^{(k)}[\boldsymbol{W}^{(k)} \boldsymbol{x}_i \| \boldsymbol{W}^{(k)} \boldsymbol{x}_p]) })  },
\end{gathered}$}
\end{equation}
where $\boldsymbol{W}^{(k)} \in \mathbb{R}^{F^{\prime} \times F}$ represents the $k$-th head learnable weight matrix which is applied to every input node feature to learn embeddings, $\boldsymbol{a}^{k}$ represents the learnable weight vector of $k$-th head, $\|$ is the concatenation operation, and 
${\rm LeakyReLU}(\cdot)$ is a nonlinear activation. Note that $\alpha_{ij}^{(k)} = 0$ if $A_{ij} = 0$.

The edge coefficients are used to compute a linear combination of the features of neighbor nodes to learn the embedding of each node.
Then, the GAT encoder uses a nonlinear activation ELU to serve as the final embeddings, as

\begin{equation}
	\boldsymbol{h}_i^{(k)} = \operatorname{ELU}(\sum_{v_j \in \mathcal{N}_i \cup \{v_i\}} \alpha_{ij}^{(k)} \boldsymbol{W}^{(k)} \boldsymbol{x}_j),
\end{equation}
where $\boldsymbol{h}_i^{(k)}$ is the embedding of node $i$ in the $k$-th head. Then, node $i$ embedding of each head is concatenated, as

\begin{equation}
	\boldsymbol{h}_i = \|_{k = 1}^{K} \boldsymbol{h}_i^{(k)}.
\end{equation}

To avoid an expensive computational burden, we exploit the node representation to generate edge representation, as

\begin{equation}
	\boldsymbol{h}_{ij} = g(\boldsymbol{e}^{ij}) = \boldsymbol{\varphi}(f(v_i),f(v_j)) =  \boldsymbol{\varphi}(\boldsymbol{h}_i, \boldsymbol{h}_j),
\end{equation}
where $f(\cdot)$ is a GNN encoder, $g(\cdot)$ represents the final learned edge representation module, and $\boldsymbol{\varphi}(\cdot,\cdot)$
is a mapping function of $\mathbb{R}^{KF^{\prime}} \times \mathbb{R}^{KF^{\prime}} \rightarrow \mathbb{R}^{D^{\prime}}$, a simple and feasible implementation as

\begin{equation}
	\boldsymbol{\varphi}(f(v_i),f(v_j)) = \boldsymbol{W}(f(v_i) \| f(v_j)) =  \boldsymbol{\varphi}(\boldsymbol{h}_i \| \boldsymbol{h}_j),
\end{equation}
where $\boldsymbol{W} \in \mathbb{R}^{D^{\prime} \times 2KF^{\prime}}$ can be an identity matrix $\boldsymbol{I}$ if $D^{\prime} = 2KF^{\prime}$
or a learnable weight matrix.

Compared to existing edge representation methods, the proposed model has two advantages.
 
$1)$ The existing edge representation learning methods usually propose to additionally create the adjacency matrix for edges, either by defining the
neighborhood structure or using the line graph transformation. {\bf However, for a nearly complete graph obtaining the adjacency of edges requires
$\mathcal{O}(n^2)$ time complexity}. We propose a simple method to generate the representation of edges, by using node representation, which can directly
avoid learning each edge representation. 

$2)$ Learning edge representations by creating a structure for edges may not yield good performance for downstream tasks, such as node classification. 
In this case, there is little correlation between the representation of nodes and the representation of edges. In contrast, our proposed method uses the
representation of {\bf two nodes connected by edges to generate the representation of edges}.

\subsection{Edge Contrastive Learning}

After obtaining the embeddings of edges, we find it necessary to define positives and negatives in edge-edge GCL.

Let $\boldsymbol{h}_i$ and $\boldsymbol{h}_j$ denote the embeddings of $v_i$ and $v_j$ learned by the GNN encoder respectively. Then select 
$\boldsymbol{h}_{ij}$ as the anchor, which is generated by $\boldsymbol{h}_i$ and $\boldsymbol{h}_j$. In our proposed model, positive samples consist of 
three sources: 
\begin{itemize}
\item the same edge, i.e., the embedding of the same edge in the view $\boldsymbol{h}_{ij}$; 
\item $\left\{\boldsymbol{h}_{ik} \mid v_k \in \mathcal{N}_i\right\}$, the embedding of the edges adjacent to node $i$;
\item $\left\{\boldsymbol{h}_{kj} \mid v_k \in \mathcal{N}_j\right\}$, the embedding of the edges adjacent to node $j$.
\end{itemize}
In this case, the number of
positive pairs associated with the anchor $\boldsymbol{h}_{ij}$ should be $|\mathcal{N}_i| + |\mathcal{N}_j| + 1$, where $|\mathcal{N}_i|$ is the number 
of neighbors of node $i$ and  $|\mathcal{N}_j|$ is the number of neighbors of node $j$. In the view, the edge contrastive loss associated with the anchor
$\boldsymbol{h}_{ij}$ is formulated as
\begin{equation}
\resizebox{1.01\linewidth}{!}{$\begin{gathered}
\displaystyle
\ell\left(\boldsymbol{h}_{ij}\right)= \\
-\log \frac{\left(\exp({\theta\left(\boldsymbol{h}_{ij}, \boldsymbol{h}_{ij}\right) / \tau})+
\sum_{v_k \in \mathcal{N}_i}\left(\exp({\theta\left(\boldsymbol{h}_{ij}, \boldsymbol{h}_{ik}\right) / \tau})\right)+
\sum_{v_k \in \mathcal{N}_j}\left(\exp({\theta\left(\boldsymbol{h}_{ij}, \boldsymbol{h}_{kj}\right) / \tau})\right)\right) 
/\left(\left|\mathcal{N}_i\right|+\left|\mathcal{N}_j\right|+1\right)}
{\exp({\theta\left(\boldsymbol{h}_{ij}, \boldsymbol{h}_{ij}\right) / \tau})+
\sum_{k \neq i \neq j}\left(\exp({\theta\left(\boldsymbol{h}_{ij}, \boldsymbol{h}_{ik}\right) / \tau})\right)+
\sum_{k \neq j \neq i}\left(\exp({\theta\left(\boldsymbol{h}_{ij}, \boldsymbol{h}_{kj}\right) / \tau})\right)},
\end{gathered}$}
\end{equation}
where $\tau$  is a temperature parameter and $\theta(\cdot)$ is the cosine similarity measure. We decompose the last two terms of the denominator of 
Eq. (6) as
$$
\resizebox{0.9\linewidth}{!}{$\begin{aligned}
& \sum_{k \neq i \neq j} \exp({\theta\left(\boldsymbol{h}_{ij}, \boldsymbol{h}_{ik}\right) / \tau})=\sum_{v_k \in \mathcal{N}_i} \exp({\theta\left(\boldsymbol{h}_{ij}, \boldsymbol{h}_{ik}\right) / \tau})+\sum_{v_k \notin \mathcal{N}_i} \exp({\theta\left(\boldsymbol{h}_{ij}, \boldsymbol{h}_{ik}\right) / \tau}), \\
& \sum_{k \neq j \neq i} \exp({\theta\left(\boldsymbol{h}_{ij}, \boldsymbol{h}_{kj}\right) / \tau})=\sum_{v_k \in \mathcal{N}_j} \exp({\theta\left(\boldsymbol{h}_{ij}, \boldsymbol{h}_{kj}\right) / \tau})+\sum_{v_k \notin \mathcal{N}_j} \exp({\theta\left(\boldsymbol{h}_{ij}, \boldsymbol{h}_{kj}\right) / \tau}),
\end{aligned}$}
$$
where the edges connected by the non-neighbor nodes of node $i$ and the edges connected by the non-neighbor nodes of node $j$ are regarded as negative pairs respectively. 
\begin{algorithm}[!h]
\caption{The pseudo-code for the proposed AFECL}
    \label{alg:algorithm1}
\renewcommand{\algorithmicrequire}{\textbf{Input:}}
\renewcommand{\algorithmicensure}{\textbf{Output:}}
    \begin{algorithmic}[1] 
\REQUIRE The feature matrix $\boldsymbol{X}$, adjacency matrix $\boldsymbol{A}$, edge feature matrix $\boldsymbol{E}$, Edge sampling probability $p_s$
        \ENSURE Embeddings of nodes $\boldsymbol{H}$    
\FOR{epoch in 1 to $\mathcal{T}$}
\FOR{$k$ in 1 to $K$}
\STATE Calculate the edge coefficient $\alpha_{ij}^{(k)}$ by Eq.(1).
\STATE Generate node embeddings $\boldsymbol{h}_i^{(k)}$ by Eq.(2).
\ENDFOR
\STATE Compute node embedding by concatenation $\boldsymbol{h}_i = \|_{k = 1}^{K} \boldsymbol{h}_i^{(k)}$.
\IF {$p_s \neq 1$}
   \STATE Sampling the edges $\boldsymbol{A}^{\prime} = \boldsymbol{A} \circ \boldsymbol{R}$.
\STATE Compute edge embeddings by Eqs.(4), (5).

\ELSE
\STATE Generate embeddings of edges by Eqs.(4), (5).
\ENDIF
\STATE Compute edge contrastive loss $\mathcal{L}$ by Eq.(7).
\STATE Update parameters to minimize $\mathcal{L}$;
\ENDFOR
    \end{algorithmic}
\end{algorithm}
Minimizing Eq. (6) is equivalent to maximizing the agreement between positive pairs and minimizing that of negative pairs. That is, the embedding of each edge is driven to agree with itself and the embeddings of the edges connected to its neighboring nodes within a view. In the meanwhile,
the embedding of each edge would disagree with the embedding of edges connected to non-neighbor nodes. Note that our proposed model is different from the existing multi-view graph contrastive learning model. In our model, positive samples include the anchor itself to prevent the learned representation from deviating from our objective. Besides, we do not need to set up a special asymmetric network structure or data augmentations to achieve contrastive learning. The final edge contrastive loss within a view, averaged over all edges, is defined as
\begin{equation}
{\mathcal{L}}=\frac{1}{M}\sum_{k=1}^M\ell\left(\boldsymbol{h}_{e_{k}}\right) = 
\frac{1}{M}\sum_{i=1}^N\sum_{j=1}^N\ell\left(\boldsymbol{h}_{ij}\right),
\end{equation}
where $\ell\left(\boldsymbol{h}_{ij}\right)=0$ if $A_{ij}=0$.

\noindent{\bf{Edge Sampling.}} We randomly sample a portion of the edges in the original graph. Since the number of edges is much more than the number of nodes, in large graphs we need to sample edges to achieve edge contrastive learning. Specifically, for simplicity, we first sample a random masking matrix 
$\boldsymbol{R}\in\{0,1\}^{N \times N}$, each entry of which follows a Bernoulli distribution $\boldsymbol{R}_{ij} \sim {\textit{Bernoulli}}\left(p_s\right)$ if $A_{ij} = 1$ and $\boldsymbol{R}_{ij} = 0$ otherwise. Here $p_s$ represents the probability of each edge being sampled.
In this way, the final sampled adjacency matrix can be computed as  
\begin{equation}
	\boldsymbol{A}^{\prime} = \boldsymbol{A} \circ \boldsymbol{R},
\end{equation}
where $\circ$ is the Hadamard product.

To sum up, at each training epoch, our model first passes through the GNN encoder to obtain the embeddings of the nodes. Then, we perform edge sampling on large graphs that do not require this operation on small graphs, and generate embeddings of these edges that are concatenated by connecting node embeddings. Finally, the parameters are updated by minimizing the objective in Eq. (7). The optimization process of our model is summarized in Algorithm~\ref{alg:algorithm1}.

\noindent{\bf{Computational Complexity.}} The original graph is fed into a multi-head GAT encoder to learn the representation of nodes, where the time complexity is 
$\mathcal{O}\left( (NFF^{\prime} + MF^{\prime})K \right)$. Note that $N$ and $M$ are the number of nodes and edges in the graph $\mathcal{G}$ respectively, $K$ is the number of heads, $F$ is the number of input node features, and $F^{\prime}$ is the node embedding dimension. Edge representation learning is simply generated by concatenating the learned corresponding node embeddings, so the time complexity can be ignored. For small graph datasets, the time complexity of edge contrastive learning is $\mathcal{O}(M^{2}D^{\prime})$. For large graph datasets, the time complexity of edge contrastive learning is $\mathcal{O}({M^{\prime}}^{2}D^{\prime})$. In both cases, $M$ or $M^{\prime}$ is slightly larger than $N$ and $D^{\prime} = KF^{\prime}$, so $\mathcal{O}(M^{2}D^{\prime})$ or $\mathcal{O}({M^{\prime}}^{2}D^{\prime})$ can be considered to be equivalent to $\mathcal{O}(N^{2}KF^{\prime})$. Thus, the time complexity of AFECL is $\mathcal{O}\left( (NFF^{\prime} + MF^{\prime})K + N^{2}KF^{\prime} \right)$. Since $M \ll N^{2}$, the overall time complexity of AFECL is $\mathcal{O}\left( (NFF^{\prime}+ N^{2}F^{\prime})K \right)$. So the time complexity of edge-edge AFECL is comparable to representative node-node GCL methods, e.g., GRACE \cite{zhu2020deep}.

\section{Experiments}

\subsection{Datasets}

In our experiments, there are totally eight benchmark datasets of node classification, which have been widely used in previous GCL
methods. In the homophilic graph datasets, three citation networks include Cora, Citeseer, and Pubmed \cite{sen2008collective}, a co-author network includes Coauthor-CS \cite{shchur2018pitfalls} and a co-purchase network includes Amazon-photo \cite{shchur2018pitfalls}. For heterophilic graphs, we adopt Actor, Chameleon, and a larger graph, Penn94. The details of these datasets can be found in Appendix A.

\subsection{Baselines}

We consider 12 state-of-the-art methods for comparison on semi-supervised node classification task. Baselines trained without labels:
CCA-SSG \cite{zhang2021canonical}, DGI \cite{veličković2018deep}, BGRL \cite{thakoor2021large}, MVGRL \cite{hassani2020contrastive}, GRACE \cite{zhu2020deep}, GCA \cite{zhu2021graph},
ARIEL \cite{feng2022adversarial}, SPGCL \cite{wang2022single}, GraphACL \cite{xiao2024simple}, and PiGCL \cite{he2024new}. Baselines trained with labels: GCN \cite{kipf2016semi}, GAT \cite{velivckovic2017graph}.

\subsection{Experimental Settings}

The proposed model was implemented using PyTorch 1.13.1 \cite{paszke2019pytorch} and Deep Graph Library 1.1.2 \cite{wang2019deep}, and trained by the Adam optimizer on all datasets. The detailed hyperparameters are in Appendix A.

We test AFECL on both semi-classification and link prediction.
In scenarios with extremely limited labels, where the number of training nodes per class $c$ is selected from {1, 2, 3, 4}, we conducted experiments following \cite{shen2023neighbor}. Besides, we also followed \cite{li2018deeper,li2019label} to refrain from utilizing a validation set with additional labels for model selection. To better verify whether the proposed method works, we conducted experiments with relatively sufficient labels. Specifically, we randomly select 20 training nodes per class. For citation networks, we followed \cite{yang2016revisiting}, which selects 500 nodes per class for validation and the rest of the nodes for testing. For a large graph dataset Penn94, we follow \cite{yang2024graphcontrastivelearningheterophily} to {\bf verify the scalability of AFECL}. Further detailed introduction can be found in Appendix A. For other graph networks, we followed \cite{liu2020towards}, which selects 30 nodes per class for validation and the rest of the nodes for testing. 

\begin{table*}[]

\renewcommand\arraystretch{1}
\centering
    \begin{adjustbox}{width=1\textwidth} 
\begin{tabular}{ccccccccccccccc}
\toprule
\multirow{2}{*}{Datasets}                                                      & \multirow{2}{*}{$c$} & \multicolumn{13}{c}{Methods}                                                                                                                          \\ \cline{3-15} 
                                                                               &                    & GCN       & GAT       & CCA-SSG      & DGI       & BGRL       & MVGRL     & GRACE    & GCA       & ARIEL     & SPGCL     & GraphACL     & PiGCL     & Ours               \\ \hline
\multirow{5}{*}{Cora}                                                          & 1                  & 42.6±11.6 & 42.1±9.5  & 53.3±10.5 & 55.4±11.4 & 52.3±10.0  & \ul{59.1±10.9} & 51.0±9.8 & 58.4±10.9 & 56.3±9.3  & 50.3±8.8 & 54.7±12.0 & 53.2±11.5 & \textbf{63.6±10.9} \\
                                                                               & 2                  & 55.0±7.5  & 53.2±9.0  & 61.2±8.5  & 64.9±9.0  & 62.4±7.3  & \ul{67.8±8.6}  & 59.7±7.9 & 66.0±7.8  & 65.8±8.0  & 62.0±6.5  & 64.1±9.0  & 61.8±9.2 & \textbf{72.4±7.7}           \\ 
                                                                               & 3                  & 63.1±6.8  & 63.2±5.3  & 65.7±5.6  & 71.1±5.6  & 68.0± 4.2  & \ul{74.5±4.1}  & 64.0±6.6 & 71.5±4.6  & 72.0±5.6  & 69.2±4.4  & 68.8±5.5  & 68.4±6.6 & \textbf{76.4±4.7}           \\ 
                                                                               & 4                  & 66.4±6.4  & 66.3±5.9  & 68.4±4.6  & 72.9±4.5  & 70.2±4.7  & \ul{76.1±3.2}  & 66.1±5.4 & 72.9±4.3  & 74.6±4.8  & 71.7±4.5  & 70.9±5.0  & 71.3±4.8 & \textbf{78.0±3.3}           \\ 
                                                                               & 20                 & 79.6±1.8  & 81.2±1.6  & 81.4±1.6  & 82.1±1.3  & 80.7±1.4  & \textbf{82.4±1.5}  & 79.6±1.4 & 79.0±1.4  & 81.3±1.3  & 81.3±1.3  & 82.0±1.1  & 80.0±1.5 & \ul{ 82.1±1.3}     \\ \hline
\multirow{5}{*}{\begin{tabular}[c]{@{}c@{}}Cite\\ seer\end{tabular}}           & 1                  & 33.8±5.9  & 31.0±7.2  & 42.5±6.5 & 47.2±9.2  & 39.5±6.0  & 32.8±8.4  & 40.3±7.2 & 38.7±9.0  & \ul{48.7±10.5} & 42.5±5.6 & 44.0±6.2  & 41.9±9.7 & \textbf{50.9±12.8}          \\ 
                                                                               & 2                  & 44.8±5.5  & 41.1±7.2  & 53.9±5.0  & \ul{58.6±4.3}  & 47.9±5.2  & 47.8±7.5  & 48.5±6.0 & 49.6±5.3  & 57.3±4.8  & 53.0±4.7  & 54.2±4.1  & 56.7±5.5 & \textbf{62.6±5.9}           \\ 
                                                                               & 3                  & 49.2±5.1  & 48.6±6.7  & 59.1±3.9  & \ul{63.3±4.3}  & 51.7±4.9  & 55.2±6.7  & 52.7±4.6 & 54.2±4.7  & 61.8±3.1  & 59.0±3.7  & 59.3±4.4  & 61.9±3.8 & \textbf{66.7±2.9}           \\ 
                                                                               & 4                  & 51.7±4.5  & 52.8±6.6  & 62.7±2.2  & \ul{65.8±2.1}  & 54.1±3.4  & 59.3±5.5  & 56.0±3.9 & 57.3±3.3  & 64.4±2.2  & 61.8±2.7  & 62.6±2.5  & 64.7±3.6 & \textbf{67.6±2.9}           \\ 
                                                                               & 20                 & 66.0±1.2  & 68.9±1.8  & 71.5±1.6  & \textbf{71.6±1.2}  & 66.0±1.6  & 71.1±1.4  & 67.0±1.7 & 65.6±2.4  & 70.9±1.4  & 69.7±1.2  & 71.5±1.4  & 71.2±1.1 & \ul{71.3±1.3}           \\ \hline
\multirow{5}{*}{\begin{tabular}[c]{@{}c@{}}Pub-\\ Med\end{tabular}}            & 1                  & 48.6±7.1  & 47.9±8.5  & 48.5±8.9 & 50.0±9.5  & 53.1±10.5 & 55.3±9.3  & 46.5±7.0 & \ul{57.7±10.5} & 49.4±7.7  & 49.6±7.7  & 50.3±8.0  & 50.2±6.4 & \textbf{59.8±12.6}          \\ 
                                                                               & 2                  & 55.8±7.1  & 54.5±7.7  & 57.2±6.5 & 58.5±8.7  & 60.1±6.8  & 62.7±7.0  & 53.8±6.9 & \ul{66.3±7.6}  & 55.6±5.5  & 56.7±6.8  & 55.6±6.9  & 54.9±8.2 & \textbf{66.6±9.4}           \\  
                                                                               & 3                  & 62.1±7.3  & 61.5±6.8  & 62.4±6.4 & 62.4±7.2  & 64.9±6.0  & 68.5±5.8  & 55.6±7.9 & \textbf{71.9±5.4}  & 59.3±5.7  & 60.5±5.0  & 60.6±6.9  & 57.3±9.1 & \ul{71.3±7.1}           \\  
                                                                               & 4                  & 65.1±5.9  & 64.2±6.1  & 65.7±6.2 & 64.1±6.2  & 67.2±5.2  & 70.6±6.0  & 57.7±6.8 & \textbf{73.6±5.4}  & 60.9±6.1  & 63.7±4.0  & 64.2±6.3  & 60.4±5.9 & \ul{72.6±5.7}           \\ 
                                                                               & 20                 & 79.0±2.5  & 78.5±1.8  & 78.8±1.8  & 78.3±2.4  & 78.1±2.6  & 79.5±2.2  & 74.6±3.5 & \textbf{81.5±2.5}  & 74.2±2.5  & 74.9±2.4  & 78.6±1.9  & 76.5±3.5 & \ul{81.2±1.7}           \\ \hline
\multirow{5}{*}{\begin{tabular}[c]{@{}c@{}}Co-\\ au-\\ thor\\ CS\end{tabular}} & 1                  & 64.8±8.8  & 64.2±9.0  & 68.4±9.1  & 71.4±6.3  & \ul{78.0±5.7}  & 75.4±7.2  & 60.0±7.7 & 59.9±7.6  & 75.1±7.2  & 68.7±6.1  & 50.5±3.8  & 43.0±11.5 & \textbf{78.9±6.7}           \\  
                                                                               & 2                  & 79.2±4.2  & 80.2±4.1  & 77.4±5.0  & 79.6±5.3  & \ul{84.9±3.0}  & 84.7±2.7  & 71.3±4.5 & 72.5±4.6  & 83.7±3.6  & 80.0±3.3  & 61.5±3.2  & 62.7±8.9 & \textbf{85.5±4.3}           \\ 
                                                                               & 3                  & 83.3±4.0  & 85.0±2.7  & 80.8±4.1  & 82.3±3.6  & 86.6±2.4  & \textbf{87.5±2.2}  & 74.8±3.8 & 77.9±4.1  & 86.2±2.6  & 79.5±4.5  & 67.3±3.1  & 72.0±5.6 & \ul{87.0±3.4}           \\ 
                                                                               & 4                  & 84.2±3.1  & 86.6±2.1  & 81.5±3.5  & 84.8±2.8  & 87.4±1.7  & \textbf{88.5±1.8}  & 77.6±2.8 & 80.3±3.1  & 87.0±1.8  & 81.6±3.0  & 71.9±2.8  & 77.2±4.5 & \ul{87.9±2.7}           \\ 
                                                                               & 20                 & 90.0±0.6  & 90.9±0.7  & 88.3±1.4  & \textbf{92.0±0.5}  & 90.4±0.5  & 91.5±0.6  & 90.0±0.7 & 90.9±1.1  & 90.2±0.9  & 87.9±1.1  & 86.9±1.2  & 91.0±0.7 & 90.9±1.3           \\ \hline
\multirow{5}{*}{\begin{tabular}[c]{@{}c@{}}Ama-\\ zon\\ Photo\end{tabular}}    & 1                  & 60.7±9.3  & 59.0±11.5 & 67.0±9.1  & 53.8±10.7 & 68.2±6.6  & 59.7±9.0  & 67.0±9.0 & 55.3±6.7  & \ul{69.0±7.8}  & 53.5±4.6  & 65.3±9.1  & 21.9±6.1& \textbf{73.9±6.7}           \\ 
                                                                               & 2                  & 75.2±7.2  & 71.7±6.4  & 76.4±4.6  & 62.7±8.5  & \ul{78.4±5.3}  & 73.4±6.8  & 76.6±5.2 & 68.0±5.6  & 75.5±7.1  & 67.4±4.7  & 74.4±5.4  & 31.3±8.5 & \textbf{81.3±2.6}           \\ 
                                                                               & 3                  & 76.9±5.1  & 75.6±6.3  & 79.6±2.9  & 66.6±7.7  & \ul{80.7±4.5}  & 76.8±6.1  & 78.6±4.8 & 74.4±5.9  & 78.0±5.2  & 73.1±5.4  & 79.3±4.8  & 37.4±7.7 & \textbf{82.7±3.0}           \\ 
                                                                               & 4                  & 81.0±4.6  & 79.3±5.9  & 81.5±2.5  & 70.8±6.0  & \ul{82.8±3.6}  & 82.0±2.3  & 81.8±1.4 & 78.8±3.9  & 80.9±4.6  & 77.1±3.6  & 81.2±5.1  & 43.5±6.0 & \textbf{84.0±2.9}           \\ 
                                                                               & 20                 & 86.3±1.6  & 86.5±2.1  & 89.0±1.4  & 83.5±1.2  & 89.1±1.1  & \ul{89.7±1.2}  & 87.9±1.4 & 87.0±1.9  & \textbf{90.6±1.8}  & 88.8±1.4  & 90.0±1.0  & 71.8±3.4 & 89.2±1.2           \\ \bottomrule
\end{tabular}
\end{adjustbox}
\caption{Node classification performance with different label rates on homophilic graph datasets.}
\label{tab:results}
\end{table*}

\begin{table}[]
 \centering
    \begin{adjustbox}{width=0.47\textwidth} 
\begin{tabular}{cccccccc}
\toprule
Methods                  & $c$  & CCA-SSG & BGRL & SP-GCL & GraphACL   & PiGCL & AFECL\\ \hline
\multirow{5}{*}{Actor}     & 1  & 20.3±1.6 & 20.6±1.8     & 20.8±1.9   & 19.6±2.4 & 19.6±3.2 & \textbf{21.9±3.0} \\  
                         & 2  & 21.7±1.6 & 21.0±1.3     & 22.1±1.7   & 21.2±2.5 & 19.9±3.2 & \textbf{22.1±3.7} \\  
                         & 3  & 21.8±1.1 & 21.2±1.3     & 22.3±1.9   & 21.3±2.4 & 20.7±2.9 & \textbf{22.5±3.1} \\ 
                         & 4  & 22.4±1.7 & 21.4±1.0     & 22.4±2.0   & 21.5±1.8 & 21.1±1.9 & \textbf{22.6±3.2} \\  
                         & 20 & 24.6±1.5 & 21.5±1.2     & 23.8±1.2   & 22.1±1.5 & 21.1±1.7 & \textbf{25.0±2.3}\\ \hline
\multirow{5}{*}{Chameleon} & 1  & 22.9±3.6 & 23.6±3.7     & 24.6±3.5   & 25.8±2.7 & 21.1±4.0 & \textbf{26.9±4.0} \\  
                         & 2  & 27.4±4.6 & 27.3±2.5     & 27.7±3.5   & 28.5±4.2 & 25.2±3.4 & \textbf{30.1±3.4} \\ 
                         & 3  & 29.1±4.0 & 29.8±2.6     & 30.3±3.1   & 29.3±3.0 & 26.4±4.4 & \textbf{30.8±3.8} \\  
                         & 4  & 31.9±3.5 & 30.5±2.7     & 32.0±3.7   & 32.0±2.6 & 26.9±2.8 & \textbf{33.0±3.7} \\ 
                         & 20 & 46.3±2.8 & 35.3±1.7     & 48.1±3.3   & 49.2±2.8 & 34.1±2.6 & \textbf{49.5±3.0} \\ \bottomrule
\end{tabular}
\end{adjustbox}
\caption{Node classification performance with different label rates on heterophilic graph datasets.}
\label{tab:hete}
\vspace{-2mm}
\end{table}

Each self-supervised GCL model is trained in an unsupervised manner and tests a simple L2-regularized logistic regression classifier for semi-supervised node classification. Note that for the self-supervised GCL models to learn embedding without labels, the labeled validation set is just used to tune the hyperparameters of the logistic regression classifier. We trained the model 20 runs for different random splits of data and reported the averaged performance of all models on each dataset. For all baselines, we use the official code published by the authors. We also report previously published results of other methods as done in \cite{shen2023neighbor}. As for link prediction, we followed \cite{shiao2023link} in transductive settings. Specifically, we first train the encoder by the different models and freeze its weight before training the decoder to evaluate
the link prediction task. We run this 5 times and report the result of the mean average.

\subsection{Overall Performance}

{\bf{Node Classification.}} The results of homophilic and heterophilic graphs node classification accuracy are summarized in Table~\ref{tab:results} and 
From the tables, it is not hard to find that the proposed model shows strong performance across five benchmark graph datasets. 
Table~\ref{tab:hete}.

Specifically, the proposed model achieves the best 
or second-best results when the labels are deficient (i.e., only 1, 2, 3 and 4 labeled training nodes per class). 
\begin{table}[h]
    \centering
    \begin{adjustbox}{width=0.47\textwidth} 
\begin{tabular}{cccccc}
\toprule
Dataset     & E2E-GCN            & CCA-SSG        & BGRL       & T-BGRL        & AFECL       \\ \hline
Cora        & 91.1±0.4  & 64.7±7.6 & 91.1±0.8 & 91.0±0.5 & \textbf{96.5±0.0} \\ \hline
Citeseer    & 92.2±0.6  & 66.1±5.0 & 93.4±0.9  & 95.3±0.3 & \textbf{96.6±0.2} \\ \hline
Coauthor-CS & 96.4±0.5 & 75.8±4.7  & 95.9±0.2 & 95.6±0.2 & \textbf{98.2±0.2} \\ \bottomrule
\end{tabular}}

\end{adjustbox}
\caption{Area under the ROC curve for the methods in link prediction.
 }
\label{tab:link}
\vspace{-2mm}
\end{table}
For relatively sufficient labels, i.e., 20 labeled training nodes per class, our proposed model is competitive with the previous state-of-the-art methods. The strong performance verifies the superiority of the proposed model.
The performance of the proposed model is mainly attributed to two folds. 

Firstly, general augmentation (node masking, dropping edges and etc.) adopted in GCL baselines may alter the semantics of datasets and lead to ineffective embeddings. Our proposed model does not need to use augmentation, which avoids destroying the original topology. The existing baselines tend to regard edges as auxiliary information to learn the representation of nodes. However, the number of edges is significantly more than the number of nodes, which indicates that edge information is relatively wealthy. Our proposed model utilizes edges as the contrastive samples to define positives and negatives, which is different from node-node GCL baselines. Compared to existing baselines, our proposed method performs strongly, and the model performs relatively better than other baselines when there is less labeled data.

Secondly, the existing GCL baselines generally adopt contrastive losses, which treat all other nodes except themselves as negatives without considering the graph topology information. We proposed a novel contrastive loss regarding edges connected to the same node as positives and the remaining as negatives to avoid graph augmentation. Besides, the edge contrastive loss can take full advantage of network topology owing to the definition of positives and negatives following the homogeneity assumption. For node classification, using edge contrastive loss can extract higher--level representation to improve performance when the labeled rate of nodes is low. 
In summary, the performance of our proposed model compared to existing state-of-the-art baselines verifies the effectiveness of our model. AFECL not only avoids altering the structure of data but also exploits the graph topology to learn the representations.

\noindent{\bf{Link Prediction.}} In the link prediction task, we evaluate the AFECL on the Cora, Citeseer, and Coauthor-CS datasets, as shown in Table.~\ref{tab:link}. Intuitively, our proposed model is an edge--evel contrastive method, which motivates AFECL to be effective in link prediction.

\subsection{Ablation Study}
\begin{table}[]

\centering
    \begin{adjustbox}{width=0.47\textwidth} 
\begin{tabular}{ccccccccc}
\toprule
Method                  & $c$  & Cora & CiteSeer & PubMed & CS   & Photo & Actor & Chameleon\\ \hline
\multirow{5}{*}{ECL}     & 1  & 63.6 & 50.9     & 59.8   & 78.9 & 73.9  & 21.9 & 26.9\\  
                         & 2  & 72.4 & 62.6     & 66.6   & 85.5 & 81.3  & 22.1 & 30.1\\  
                         & 3  & 76.4 & 66.7     & 71.3   & 87.0 & 82.7  & 22.5 & 30.8 \\ 
                         & 4  & 78.0 & 67.6     & 72.6   & 87.9 & 84.0  & 22.6 & 33.0\\  
                         & 20 & 82.1 & 71.3     & 81.2   & 90.9 & 89.2  & 25.0 & 49.5\\ \hline
\multirow{5}{*}{w/o ECL} & 1  & 48.1 & 41.7     & 45.1   & 50.7 & 54.2  & 20.9 & 26.4\\  
                         & 2  & 57.2 & 49.9     & 49.7   & 62.8 & 67.6  & 21.2 & 29.8\\ 
                         & 3  & 63.2 & 55.2     & 52.8   & 68.1 & 72.0  & 21.5 & 30.5\\  
                         & 4  & 66.1 & 58.1     & 55.6   & 69.6 & 75.4  & 21.6 & 32.5\\ 
                         & 20 & 76.7 & 67.0     & 64.7   & 78.6 & 83.6  & 24.8 & 48.4\\ \bottomrule
\end{tabular}
\end{adjustbox}
\caption{Ablation study on node classification with different label rates.}
\label{tab:ablation}
\end{table}
\begin{figure}
	\centering

    \subcaptionbox{\label{CS}Coauthor-CS}{
        \includegraphics[width=0.95\linewidth]{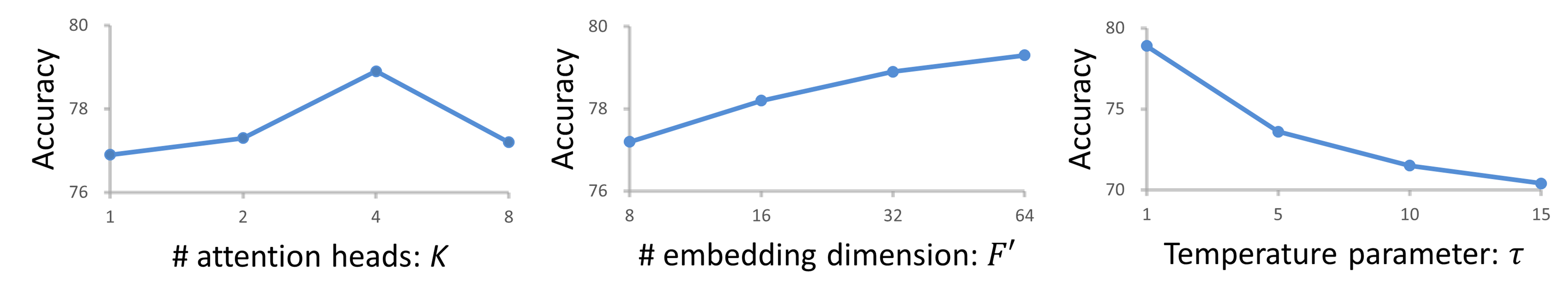}
    }
    \subcaptionbox{\label{photo}Amazon-Photo}{
        \includegraphics[width=0.95\linewidth]{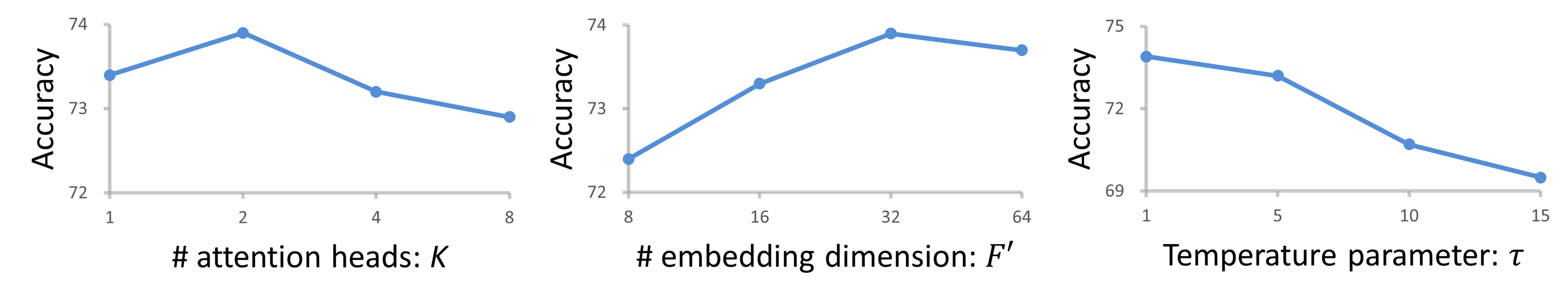}
    }

	\caption{Sensitivity analysis of the hyperparameters $K$, $F^{\prime}$ and $\tau$ on our model.}
	\label{hyperparameters1}
\vspace{-2mm}
\end{figure}

In this section, we conduct ablation experiments to demonstrate the effectiveness of edge contrastive loss of the proposed model.
\begin{figure}[htbp]
\centering
\includegraphics[scale=0.16]{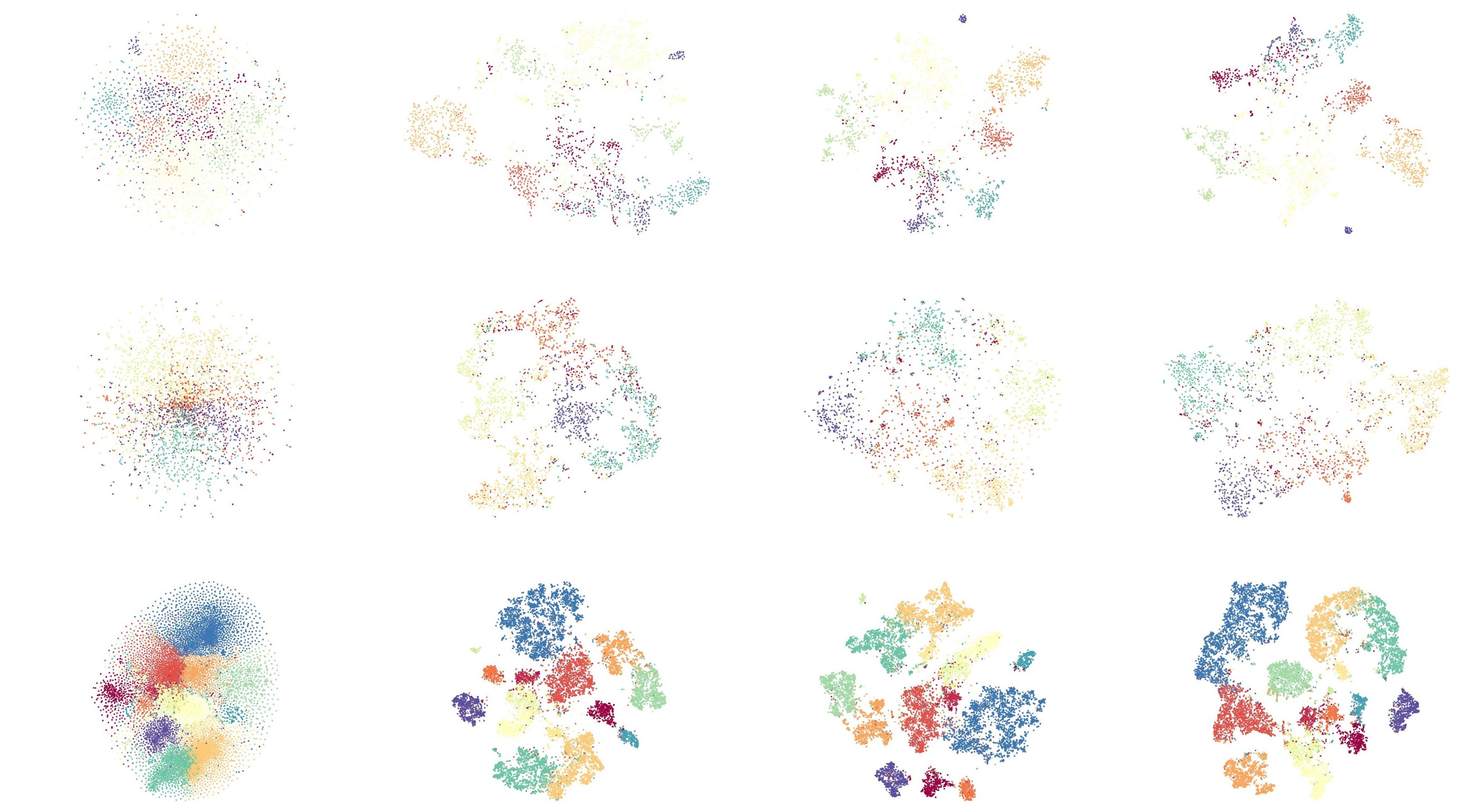}
	\caption{t-SNE visualization of representations on benchmark datasets. The rows represent the visualization results on Cora, CiteSeer, and Coauthor-CS, respectively. The columns suggest the results on the original dataset, GCA, GraphACL, and AFECL.}
	\label{visual1}
\vspace{-2mm}
\end{figure}
Table~\ref{tab:ablation} shows the classification accuracy of our method for different label rates when using different edge contrastive losses to define different positive and negative pairs.
The first loss variant ECL is our proposed method.
The second loss variant w/o ECL is to define edges corresponding to nodes connected by only one node as negative pairs.
Note that in this case w/o ECL can be regarded as defining two identical augmented graphs corresponding to the same edge as a positive pair, and the same augmented graph with different edges as a negative pair.
From the table, we can observe that the proposed edge contrastive loss consistently yields the highest accuracy in the node classification task across the five datasets. Moreover, if data augmentation is not used, defining edges corresponding to nodes connected by only one node as negative pairs will lead to poor results. This reflects that defining positive and negative pairs by considering topological information is very effective for AFECL.


\subsection{Hyperparameter Analysis}

Figure \ref{hyperparameters1} shows the sensitivity analysis on the hyperparameters $K$, $F^{\prime}$ and $\tau$ of our model. Since the size of the dimension of 
$D^{\prime} = 2KF^{\prime}$, we do not need additional sensitivity analysis of the parameter $D^{\prime}$. For large graphs, our method requires edges 
sampling, so we observe the impact of different parameters on two various datasets, i.e., Coauthor-CS(co-author network) and Amazon-photo(co-purchase network) as examples. 
We observe that more heads lead to higher complexity and experiments often achieve the best results when $K=2$ or $K=4$. Our model performs better on both two datasets with embedding dimensions $F^{\prime}$ of $\{32, 64\}$, and has the worst results on embedding dimensions $F^{\prime}$ of 8. Larger $\tau$ will result in lower accuracy for both datasets, note the temperature parameter $\tau$ in $\{1, 5, 10, 15\}$. A more detailed hyperparameter analysis can be found in Appendix A.

\subsection{Visualization of Embeddings}

To more intuitively understand the node embeddings learned by exploiting edge contrast, we use t-SNE \cite{van2008visualizing} to visualize the graph representations of AFECL and the baselines, as shown in Figure \ref{visual1}. We can observe that the node embeddings generated by AFECL as well as the baseline method are grouped according to the corresponding node labels.
However, the main difference is that AFECL captures finer-grained class information compared to GCA and GraphACL.
Specifically, cluster nodes within each label group are grouped more tightly in AFECL compared to other methods.

\section{Conclusion}

In this paper, we propose AFECL, a novel contrastive framework where the contrastive samples are edges.
Firstly, we put forward the module of edge representation learning, where edge features are computed by their connected nodes. Secondly, we managed to implement the edge-level contrast by ECL.  Extensive experimental results confirm the effectiveness of AFECL.

\bibliography{aaai25}
\newpage

\section*{A. Experiment Supplement}

\subsection*{A.1 Datasets}
We followed prior works \cite{veličković2018deep, zhu2021graph, lim2021large, xiao2024simple} and evaluated AFECL on eight widely used datasets. The statistics of the datasets are given in Table \ref{tab:datasets} :

\begin{itemize}
\item {\bf {Cora}, \bf{CiteSeer}, and {PubMed}} are citation datasets, which are the most widely used node classification benchmarks. The nodes are documents and the edges are the citation relationships between nodes. A bag of word vectors is represented as the node features and the labels represent the research field of nodes.
\item {\bf{Amazon-Photo}} is co-purchase graph from Amazon. The nodes are documents and the edges are co-purchase relationships between nodes. Product reviews are the node features and the labels represent the product category.
\item {\bf {Coauthor-CS}} is from Coauthor. The nodes are authors and the edges are co-authorship relationships between nodes. A bag of word vectors is represented as the node features and the labels represent the research domains of nodes.
\item {\bf {Chamelon}} is from Wikipeida. The nodes are web pages and the edges are hyperlinks between nodes. Informative nouns are represented as node features and the labels represent the average daily traffic of nodes.
\item {\bf {Actor}} is from film-director-actor-writer. The nodes are actors and the edges are co-occurrence relationships between nodes. The keywords of pages are represented as node features and the labels are assigned five categories based on the words.
\item {\bf {Penn94}} is from Facebook 100. The nodes are students and the edges are the relationships between nodes. The node features are major, second major/minor, dorm/house, year, and high school. Each node is labeled based on the reported gender of the user.
\end{itemize}

\subsection*{A.2 Experiments on Penn94}

To verify the scalability of AFECL, we experiment on a large heterophilic graph Penn94. 
Following \cite{lim2021large} we randomly select 50\% of nodes for training, 25\% of nodes for validation, and 25\% of nodes for testing. 
We selected HLCL \cite{yang2024graphcontrastivelearningheterophily} as a new baseline since it is the most recent self-supervised model evaluated on the Penn94 dataset, 
as shown in Table \ref{tab:table1}. From the table, we can observe that AFECL can also work well on large graphs and can achieve better performance compared to baselines.

\begin{table}[]
\centering
    \begin{adjustbox}{width=0.47\textwidth} 
\begin{tabular}{ccccc}
\toprule
Datasets     & \# Nodes & \# Edges & \# Features & \# Labels \\ \hline
Cora         & 2708     & 10556    & 1433        & 7         \\ 
CiteSeer     & 3327     & 9228     & 3703        & 6         \\ 
PubMed       & 19717    & 88651    & 500         & 3         \\ 
Coauthor-CS  & 18333    & 163788   & 6805        & 15        \\ 
Amazon-Photo & 7650     & 238162   & 745         & 8         \\ 
Actor & 7600   & 33391   & 931         & 5         \\
Chameleon & 2277   & 36101   & 2325         & 5         \\
Penn94 & 41554   & 1362229   & 5         & 2         \\
\bottomrule
\end{tabular}
\end{adjustbox}
\caption{Statistics of the datasets.}
\label{tab:datasets}
\end{table}

\begin{table}[t]
\centering
    \begin{adjustbox}{width=0.45\textwidth} 

\begin{tabular}{cccccc}
\toprule
Method       & DGI      & GRACE    & BGRL     & HLCL     & AFECL    \\ \hline
Penn94 & 62.9±0.4 & 62.4±0.4 & 58.8±0.6 & 68.1±3.5 & \textbf{68.2±3.0} \\ \bottomrule
\end{tabular}
\end{adjustbox}
\caption{Test accuracy (\%) averaged over 5 runs. 
    We choose SSL methods for comparison on Penn94.
 }
\label{tab:table1}
\end{table}
\begin{table*}[t]
\centering
    \begin{adjustbox}{width=0.97\textwidth} 
\begin{tabular}{cccccccccc}
\toprule
Datasets     & \begin{tabular}[c]{@{}c@{}}\# Attention heads \\ $K$\end{tabular} & \begin{tabular}[c]{@{}c@{}}\# Hidden \\ Layers\end{tabular} & \begin{tabular}[c]{@{}c@{}}\# Embedding Dimension \\ $F^{\prime}$\end{tabular} & $D^{\prime}$   & $p_s$    & \begin{tabular}[c]{@{}c@{}}Temperature\\ $\tau$\end{tabular} & \begin{tabular}[c]{@{}c@{}}Learning \\ Rate\end{tabular} & \begin{tabular}[c]{@{}c@{}}Weight \\ Decay\end{tabular} & \begin{tabular}[c]{@{}c@{}}\# Epochs \\ $\mathcal{T}$\end{tabular} \\ \hline
Cora         & 4                                                               & 1                                                           & 32                                                                  & 256 & 1    & 1                                                       & 1e-2                                                     & 1e-4                                                    & 2000                                                    \\ 
CiteSeer     & 4                                                               & 1                                                           & 32                                                                  & 256 & 1    & 5                                                       & 1e-2                                                     & 1e-4                                                    & 2000                                                    \\ 
PubMed       & 2                                                               & 1                                                           & 32                                                                  & 128 & 0.5  & 5                                                       & 1e-3                                                     & 5e-5                                                    & 2000                                                    \\ 
Coauthor-CS  & 4                                                               & 1                                                           & 32                                                                  & 256 & 0.27 & 1                                                       & 5e-2                                                     & 1e-4                                                    & 2000                                                    \\ 
Amazon-Photo & 2                                                               & 1                                                           & 32                                                                  & 128 & 0.18 & 1                                                       & 1e-3                                                     & 1e-4                                                    & 2000                                                    \\ 
Actor & 32                                                               & 1                                                           & 8                                                                  & 512 & 1 & 1                                                       & 5e-2                                                     & 1e-4                                                    & 2000                                                    \\ 
Chameleon & 8                                                               & 1                                                           & 32                                                                  & 512 & 1 & 1                                                       & 1e-2                                                     & 1e-4                                                    & 2000                                                    \\ 
Penn94 & 32                                                               & 1                                                           & 256                                                                  & 16384 & 0.004 & 0.2                                                       & 1e-2                                                     & 1e-4                                                    & 2000                                                    \\ 
\bottomrule
\end{tabular}
\end{adjustbox}
\caption{Hyperparameter Settings of our model}
\label{tab:hyperparameter}
\end{table*}

\begin{table}[h]
\centering
\begin{tabular}{lccc}
\toprule
Method & Cora & CiteSeer & PubMed \\
\midrule
SPGCL & 247M & 319M & 1420M \\
GraphACL & 284M & 340M & 2128M \\
AFECL & 48M & 115M & 660M \\
\bottomrule
\end{tabular}
\caption{Comparison of memory costs on three citation datasets.}
\label{tab:performance}
\end{table}

\begin{table}[h]
\centering
\begin{tabular}{lc}
\toprule
Method & Ogbn-arxiv \\
\midrule
GRACE & 71.7±0.2 \\
CCA-SSG & 72.2±0.1 \\
GraphACL & 71.7±0.2 \\
AFECL & 72.5±0.2 \\
\bottomrule
\end{tabular}
\caption{Test accuracy (\%) averaged over 5 runs on ogbn-arxiv.}
\label{tab:ogbn_arxiv}
\end{table}

\subsection*{A.3 Hyperparameter Settings and Analysis}

The overall model parameters are summarized in Table \ref{tab:hyperparameter}. Specifically, we use the following hyperparameter search range.
\begin{itemize}
\item {\bf {Number of heads for GAT:}} \{1, 2, 4, 8, 16, 32\}.
\item {\bf {Number of layers:}} \{1, 2, 3\}.
\item {\bf {Hidden dimension:}} \{8, 16, 32, 64, 128, 256\}.
\item {\bf {Temperature:}} \{0.1, 0.2, 0.5, 1, 5, 10, 15\}.
\item {\bf {Learning rate:}} \{1e-2, 5e-2, 5e-3, 1e-3\}.
\item {\bf {Weight Decay:}} \{1e-4, 5e-5, 5e-4, 0\}.
\end{itemize}

\begin{figure}
	\centering

    \subcaptionbox{\label{CS2}Coauthor-CS}{
        \includegraphics[width=0.35\linewidth]{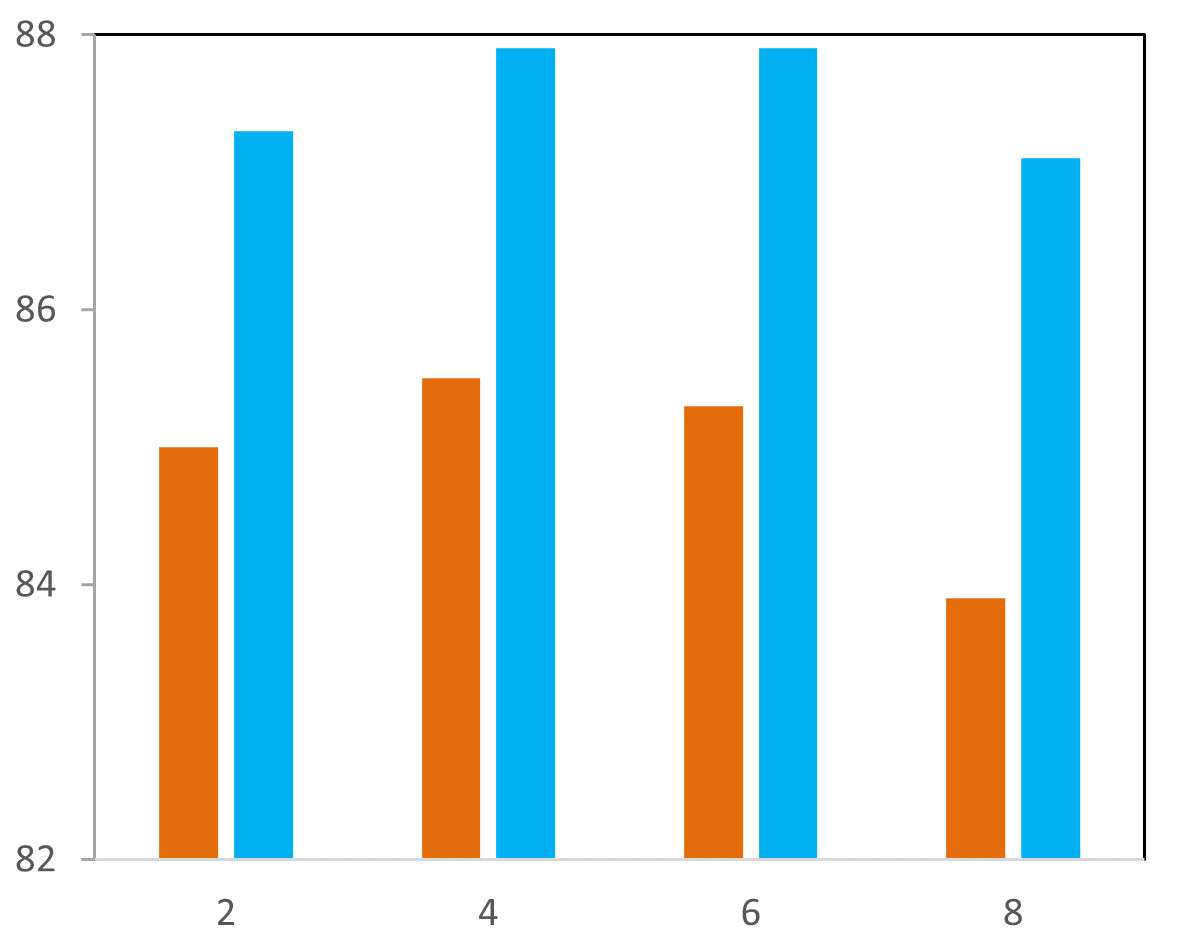}
    }
    \subcaptionbox{\label{photo2}Amazon-Photo}{
        \includegraphics[width=0.35\linewidth]{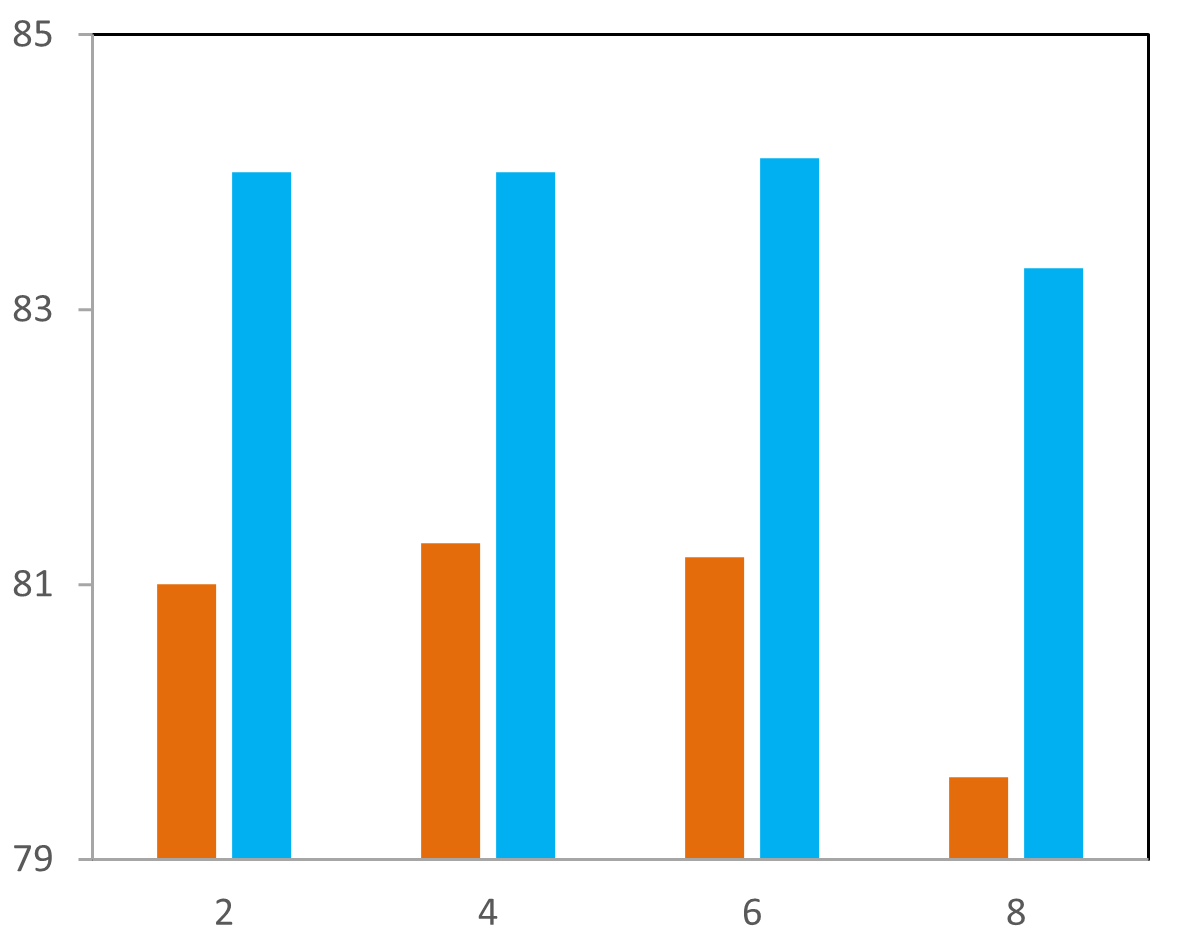}
    }

	\caption{Experimental results for edge sampling. Orange columns denote the classification accuracy for different numbers of edges when 2 labeled training nodes per class; Blue columns denote the classification accuracy for different numbers of edges when 4 labeled training nodes per class. Specifically, in the picture 6 represents 60,000 edges.}
	\label{hyperparameters2}
\end{figure}

For edge sampling probability $p_s$, the choice of this parameter is based on the number of sampled edges. In Figure \ref{hyperparameters2}, we conduct experiments on the number of edges with different sizes in $\{2, 4, 6, 8\}$ ten thousand. We observe that the performance of our model is stable when there are $40,000$ or $60,000$ edges. The recommendation to use 40,000 edges in the model is mainly attributed to two folds. On one hand, using more edges is too time-consuming during the update iteration process of the model, which is several orders of magnitude higher than 40,000 edges. On the other hand, an excessive number of edges can cause out-of-memory.

\section*{B. Efficiency and scalability Analysis}

To verify the efficiency of AFECL, we test the memory usage when the edge sampling rate is at 0.1. As shown in Table \ref{tab:performance}, AFECL requires less GPU than other augmentation-free methods.

We have conducted extra experiments on the ogbn-arxiv dataset to further validate the scalability of AFECL, as shown in Table \ref{tab:ogbn_arxiv}. Specifically, we follow StructComp \cite{zhang2024structcompsubstitutingpropagationstructural} in a single-view setting. To begin with, we use the METIS algorithm the obtain the graph partition matrix and then use a simple MLP to compute the compressed node features. The compressed edge features were then derived from the compressed node features, and the edge-level contrastive loss was applied to optimize the model.

\end{document}